\documentclass[paper=a4, fontsize=11pt]{scrartcl}
\usepackage[T1]{fontenc}
\usepackage{fourier}

\usepackage[english]{babel}															
\usepackage[protrusion=true,expansion=true]{microtype}	
\usepackage{amsmath,amsfonts,amsthm} 
\usepackage[pdftex]{graphicx}	
\usepackage{url}
\usepackage{hyperref}
\usepackage[font=small]{caption}
\usepackage{floatrow}
\usepackage{array}
\usepackage{xcolor}
\usepackage{footnote}
\usepackage{subcaption}
\usepackage{sectsty}
\allsectionsfont{\centering \normalfont\scshape}

\usepackage{fancyhdr}
\numberwithin{equation}{section}		
\numberwithin{figure}{section}			
\numberwithin{table}{section}				

\newcommand{\horrule}[1]{\rule{\linewidth}{#1}} 	

\title{
		\usefont{OT1}{bch}{b}{n}
		\normalfont \normalsize \textsc{Northeastern University} \\
		\normalfont \normalsize \textsc{Electrical and Computer Engineering Department} \\
		\normalfont \normalsize \textsc{Augmented Cognition Laboratory (ACLab)} \\
		[25pt]
		\horrule{0.5pt} \\[0.4cm]
		\large Summer 2021 Independent Study \\
		\huge Human Pose Estimation \\
		\horrule{2pt} \\[0.5cm]
}
\author{
		\normalfont 								
        By: Rohit Josyula\\[-3pt]
        \normalfont 								\normalsize
        Advisor: Prof. Sarah Ostadabbas\\[-3pt]
        \normalsize
}
\date{}

\begin{document}
\maketitle
\tableofcontents

\newpage

\section{Introduction to Human Pose Estimation (HPE)}
\begin{figure}[h]
     \centering
     \begin{subfigure}{0.4\textwidth}
         \centering
         \includegraphics[height = 6cm, width=\textwidth]{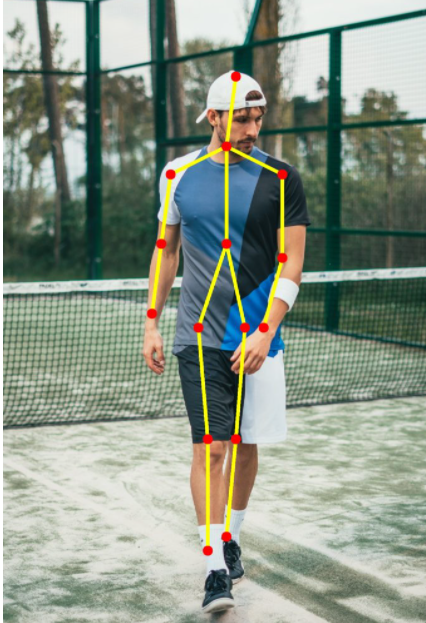}
     \end{subfigure}
     \begin{subfigure}{0.4\textwidth}
         \centering
         \includegraphics[height = 6cm, width=\textwidth]{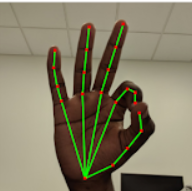}
     \end{subfigure}
     \begin{subfigure}{0.6\textwidth}
         \centering
         \includegraphics[height = 6cm, width=\textwidth]{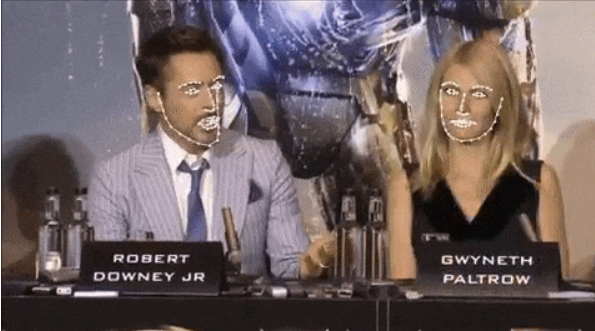}
     \end{subfigure}
        \caption{These images are examples of the different types of pose estimation. Top left image is an example of body pose estimation \cite{website:nanonets-2019}, while top right image is an example of hand pose estimation \cite{website:mp-hands}. Bottom image is an example of face pose estimation \cite{gb:adrianb}.}
    \label{fig:intro-HPE-examples}
\end{figure}

The phenomenon of Human Pose Estimation (HPE) is a problem that has been explored over the years, particularly in computer vision. But what exactly is it? To answer this, the concept of a pose must first be understood. Pose can be defined as the arrangement of human joints in a specific manner. Therefore, we can define the problem of Human Pose Estimation as the localization of human joints or predefined landmarks in images and videos \cite{website:nanonets-2019}. There are several types of pose estimation, including body, face, and hand (see Figure \ref{fig:intro-HPE-examples}), as well as many aspects to it. The rest of this paper will cover them, starting in Section \ref{sec:dp and cpm} with the classical approaches to HPE and the first Deep Learning based model. 
\vspace{1.7cm}

\vspace{18cm} 
\section{Deep Learning in HPE and What was Done Before It}
\label{sec:dp and cpm}
\vspace{0.4cm}
\begin{figure}[h]
     \centering
     \begin{subfigure}{0.4\textwidth}
         \centering
         \includegraphics[height = 3cm, width=\textwidth]{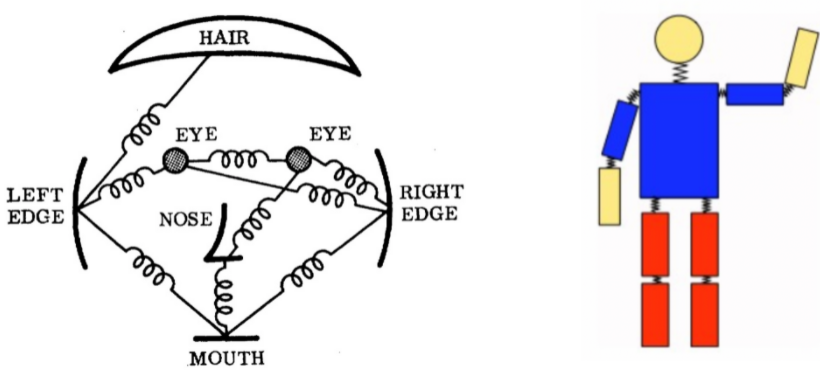}
         \caption{Pictorial Structures Model}
     \end{subfigure} 
     \hspace{1cm}
     \begin{subfigure}{0.3\textwidth}
         \centering
         \includegraphics[height = 3cm, width=\textwidth]{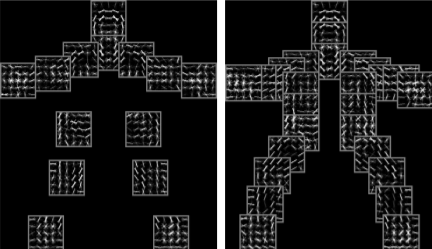}
         \caption{Flexible Mixture-of Parts}
     \end{subfigure}
        \caption{These are two examples of classical approaches to HPE. (a) Pictorial Structures Model connects rigid body parts together through the use of springs to create a tree-like structure of the entire body \cite{website:nanonets-2019}. (b) FMP utilizes a mixture of several small non-oriented parts to create a deformable configuration of the body. The picture on the left uses 14 body parts while the one on the right uses 26, demonstrating how this model uses several parts to create the pose estimation \cite{FMP}.}
    \label{fig:HPE-classics-examples}
\end{figure}
\subsection{Classical Approaches to HPE}
\begin{figure}
    \includegraphics{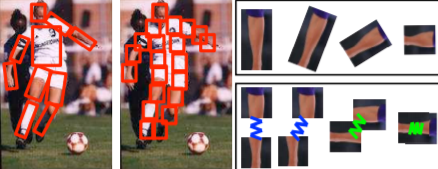}
    \caption{Shown here is a side-by-side comparison of the Pictorial Structures Model (left) and FMP (middle). Since the Pictorial Structures Model uses warped templates, there are few body parts that are oriented and big in size, as shown in the top right and left photos. FMP, however, uses more body parts that are not oriented and are smaller in size, which is why their appearance varies by location, as shown in the middle and bottom right photos \cite{FMP}.}
    \label{fig:model-comparison}
\end{figure}

Deep Learning, which is a type of machine learning that uses artificial neural networks with multiple layers to process complex data, is widely used today for HPE. However, before it was implemented, other methods were being used, as follows:
\begin{itemize}
    \item \textbf{Pictorial Structures Model} \cite{website:nanonets-2019} \cite{website:neuralet-DL-Classics} \cite{wei2016convolutional}: 
    \begin{itemize}
        \item This framework models the spatial correlations of rigid body parts by expressing them as a tree-structured graphical model in order to predict the body joints’ location. These spatial connections are shown through the use of springs, and the parts are appearance templates based on the image. By parameterizing parts using pixel location and orientation, the resulting structure can model articulation. See Figure \ref{fig:HPE-classics-examples}a for a visual representation of this model. 
    \end{itemize}
    \begin{itemize}
        \item A problem with this approach, however, is that it can’t capture correlations between invisible and deformable body parts, meaning the model is prone to errors if not all the limbs of the person are visible. It is also not dependent on image data. 
    \end{itemize}
    \item \textbf{Flexible Mixture-of Parts (FMP)} \cite{website:nanonets-2019} \cite{FMP}: 
    \begin{itemize}
        \item This approach uses the deformable part models, which are a collection of templates that are matched for in an image and are arranged in a deformable configuration. Furthermore, each model has global and part templates. The main idea is to use a mixture of small, non-oriented parts as opposed to using a family of warped, meaning rotated and foreshortened, templates. The reasoning for this has to do with the variation in how limbs appear and changes in viewpoint.
    \end{itemize}
    \begin{itemize}
        \item FMP simultaneously captures spatial relations between part locations and co-occurrence relations between part mixtures, leading to pictorial structure models that encode solely spatial relations. Through dynamic programming, the models share computation across similar warps, making this approach not only significantly fast, but also highly efficient. Additionally, they model an exponentially-large set of global mixtures through the composition of local part mixtures in order to learn notions of local-rigidity, as well as capturing the effect of global geometry on local appearance, meaning the appearance of parts varies across different locations. See Figure \ref{fig:HPE-classics-examples}b for a visual representation of this model and Figure \ref{fig:model-comparison} for a comparison between it and the Pictorial Structures Model. 
    \end{itemize}
    \begin{itemize}
        \item FMP is able to express complex joint relationships well, hence why it can model articulations well too. However, it has its problems, which include limited expressiveness and no consideration for global context.
    \end{itemize}
    \item \textbf{Edges, color histograms, contours, and histogram of oriented gradients} were other features that were applied to early works of HPE and served as the main building blocks of different classical models to determine accurate locations of body parts \cite{website:neuralet-DL-Classics}.
\end{itemize}
General problems with classical approaches include poor generalization and inaccurate body parts detection. So, Deep Learning was implemented to resolve these issues \cite{website:neuralet-DL-Classics}. 

\vspace{2cm}

\begin{figure}
    \centering
    \includegraphics[height=3cm, width=12cm]{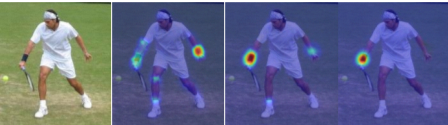}
    \caption{This is a sequential architecture across several stages using convolutional networks, spatial context from belief maps, and large receptive fields to create increasingly refined estimates for body part predictions, which in this case is the right elbow \cite{wei2016convolutional}.}
    \label{fig:CPMstages}
\end{figure}

\subsection{Deep Learning in Convolutional Pose Machine}
A pose machine consists of a sequence of multi-class predictors that are trained to predict the location of each part in each level of hierarchy. It also has an image feature computation module and a prediction module, both of which can be replaced by a convolutional architecture allowing for both image and contextual feature representations to be learned directly from data. This idea is what led to \textbf{Convolutional Pose Machine (CPM)}, which is the first Deep Learning-based pose estimation model \cite{wei2016convolutional}. 

CPM is fully differentiable, which allows its multi-stage architecture to be trained end-to-end using backpropagation, an algorithm used for training feedforward neural networks. Additionally, its sequential prediction framework that consists of convolutional networks and learns implicit spatial models utilizes larger receptive fields on the belief maps from previous stages, which assists with learning the long range spatial relationships between parts and results in improved accuracy due to increasingly refined estimates for part locations in the later stages (see Figure \ref{fig:CPMstages}) \cite{wei2016convolutional} \cite{website:review-cpm} \cite{website:nanonets-2019}. The problem of vanishing gradients, which is when backpropagated gradients diminish in strength as they are passed through many layers of the network, is addressed through intermediate supervision after each stage \cite{website:review-cpm}. 

\vspace{0.2cm}
\begin{figure}[h]
    \centering
    \includegraphics[height=5cm, width=10cm]{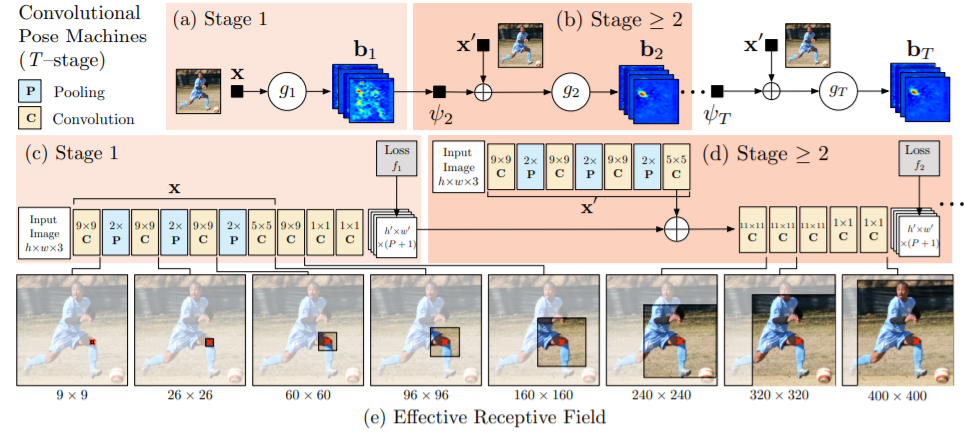}
    \caption{This picture is the architecture of a CPM with any number of stages from the paper "Convolutional Pose Machines" \cite{wei2016convolutional}. (a) and (b) show the pose machine, (c) and (d) show corresponding convolutional networks, and (e) shows the effective receptive field on the given image.}
    \label{fig:CPMarch}
\end{figure}

The first stage of CPM predicts part beliefs using just local image evidence by using a deep convolutional network consisting of 7 total convolutional layers. The belief maps created from this stage are added to the inputted data before being processed by a few convolutional layers. Towards the later stages, the effective receptive field is increased to help improve accuracy \cite{website:review-cpm}. See Figure \ref{fig:CPMarch} for an overview of the architecture of CPMs.  

Overall, this approach allows for the architecture to learn both image features and image-dependent spatial models for prediction tasks without the need for graphical model style inference \cite{wei2016convolutional}. 

\vspace{15cm}

\section{Accuracy and Metrics}
\subsection{Definition of Accuracy and Concept of Metrics}
The definition of accuracy is the evaluation of machine learning models by calculating the performance value of their algorithms \cite{website:OKS-PDJ}. There are many evaluation metrics used to do these calculations, which are described in Subsection \ref{metrics}. The reason for this is because there are many features and requirements that need to be considered when looking to evaluate the performance of a human pose estimation model \cite{zheng2021deep}. So, in other words, the accuracy of a model is defined by using metrics, meaning metrics are a way to quantify the accuracy of the models. 

\subsection{Different Metrics Used in HPE}
\label{metrics}
As stated earlier, there are several metrics used to evaluate the performance of HPE models. Listed below are a few of them:
\begin{itemize}
    \item \textbf{Intersection Over Union (IoU)} \cite{website:IOU}: This is a metric that finds the difference between ground truth annotations and predicted bounding boxes. Removes any unnecessary boxes based on the threshold value assigned, which is typically 0.5
    \item \textbf{Percentage of Correct Parts (PCP) and Percentage of Detected Joints (PDJ)} \cite{zheng2021deep} \cite{website:OKS-PDJ}: PCP is a metric that is not as commonly used now, but its purpose was to report the localization accuracy for limbs. This is determined when the distance between the predicted and ground truth joints is less than a fraction of the limb length, which is between 0.1 and 0.5. If the threshold is 0.5, the PCP measure is referred to as PCP@0.5. A higher PCP measure means better performance. A limitation to this metric, however, is that it is inaccurate for limbs with short length. Because of this, PDJ was implemented, which follows the same logic as PCP; if the distance between the predicted and true joints is within a certain fraction of the torso diameter, the joint is viewed as correctly detected. The use of this metric means the accuracy in determining all joints is based on that threshold.  
    \item \textbf{Percentage of Correct Keypoints (PCK)} \cite{zheng2021deep}: This metric is used to measure the accuracy of localization of different keypoints within a certain threshold. It is set to 50\% of the head segment length of each test image. Tying back to PDJ, when the distance between the detected and truth joints is less than 0.2 times the torso diameter, it is referred to as PCK@0.2. The higher the PCK value, the better the performance. 
    \item \textbf{Average Precision (AP)} \cite{zheng2021deep} \cite{website:MAR/MAP/AR/AP} \cite{website2:MAR/MAP/AR/AP}: AP measures the accuracy of keypoint detection according to precision, which is the ratio of true positive results to the total positive results. In other words, how accurate are the predictions. Therefore, the AP metric is the precision averages across all recall values between 0 and 1 at various IoU thresholds.
    \begin{itemize}
        \item \textbf{Mean Average Precision (MAP)} \cite{zheng2021deep} \cite{website:MAR/MAP/AR/AP} \cite{website2:MAR/MAP/AR/AP} is the mean of average precision over all classes at multiple IoU thresholds across the whole model.  
    \end{itemize}
    \item \textbf{Average Recall (AR)} \cite{zheng2021deep} \cite{website:MAR/MAP/AR/AP} \cite{website2:MAR/MAP/AR/AP}: AR measures the accuracy of keypoint detection according to recall, which is the ratio of true positive results to the total number of ground truth positives. In other words, how much of all the true positives were found by the model. Therefore, the AR metric is the recall averages across all recall values between 0 and 1 at various IoU thresholds.
    \begin{itemize}
        \item \textbf{Mean Average Recall (MAR)} \cite{zheng2021deep} \cite{website:MAR/MAP/AR/AP} \cite{website2:MAR/MAP/AR/AP} is the mean of average recall over all classes at multiple IoU thresholds across the whole model.  
    \end{itemize}
     \item \textbf{Object Keypoint Similarity (OKS)} \cite{website:OKS-PDJ} \cite{website:COCO-metrics}: This metric is the average keypoint similarity across all object keypoints. It is calculated based on the scale of the subject and the distance between predicted and ground truth points. The scale and keypoint constant is required to give equal importance to every keypoint. Every keypoint is given a similarity value from 0 to 1, and OKS is the average of all these values across all the keypoints. This metric also helps with finding AP and AR.
    \item \textbf{Mean Per Joint Position Error (MPJPE)} \cite{zheng2021deep}: This is the most widely used metric for 3D HPE. Calculated by using the Euclidean distance between the estimated 3D joints and the ground truth positions as follows:
    \large
    \begin{equation*}
        MPJPE = \frac{1}{N} \sum_{i=1}^{N} \|J_i - J_i^*\|_2, 
    \end{equation*}
    \normalsize
    where \emph{N} is the number of joints, $J_i$ and $J_i^*$ are the ground truth position and the \hspace{2 cm} estimated position of the $i_{th}$ joint, respectively.
 
\end{itemize}
\vspace{9cm}

\section{Standards and Categorization for Human Pose Estimation}
\subsection{Table of Standards}
\begin{figure}[h]
     \centering
     \begin{subfigure}{0.25\textwidth}
         \centering
         \includegraphics[height = 5cm, width=\textwidth]{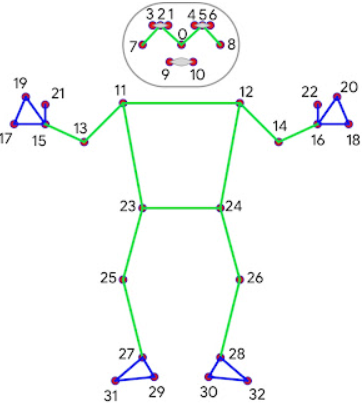}
         \caption{}
     \end{subfigure} 
     \begin{subfigure}{0.2\textwidth}
         \centering
         \includegraphics[height = 5cm, width=\textwidth]{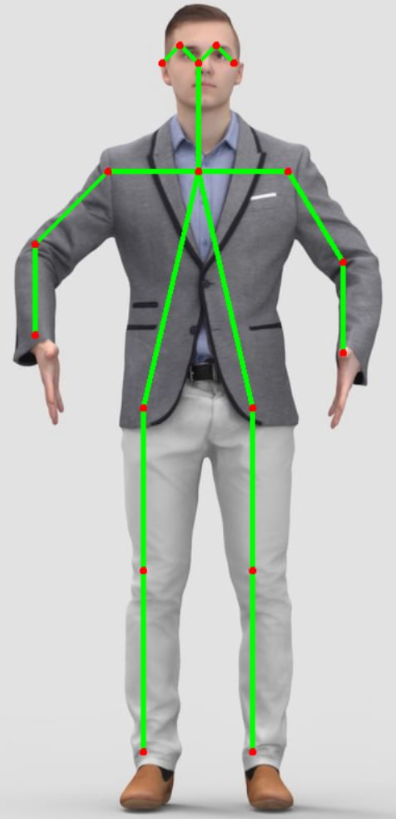}
         \caption{}
     \end{subfigure}
     \hspace{1cm}
     \begin{subfigure}{0.2\textwidth}
         \centering
         \includegraphics[height = 5cm, width=\textwidth]{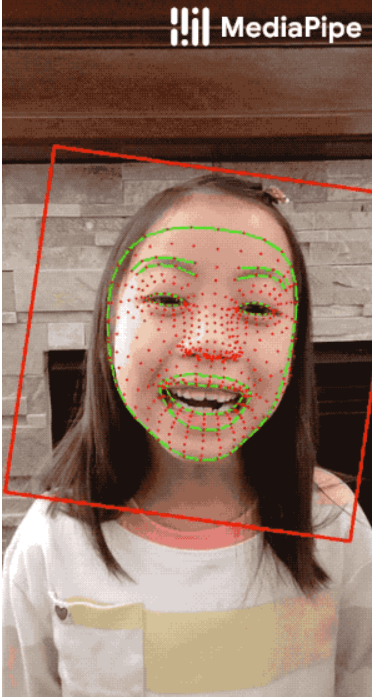}
         \caption{}
     \end{subfigure}
     \begin{subfigure}{0.25\textwidth}
         \centering
         \includegraphics[height = 5cm, width=\textwidth]{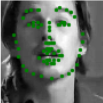}
         \caption{}
     \end{subfigure}
        \caption{These are examples of body and face pose estimation using a different number of landmarks. The body pose estimation method in (a) uses 33 landmarks \cite{website:MP-Body}, while the body pose estimation method in (b) uses 18 landmarks  \cite{gb:quanhua}. Likewise, the face pose estimation method in (c) uses 468 landmarks \cite{website:mp-facemesh}, while the face pose estimation method in (d) uses 68 landmarks \cite{gb:PBB}.}
    \label{fig:HPE-landmarks}
\end{figure}

Standards in HPE refer to the number of landmarks used to carry out the estimation, meaning the number of human joints or predefined landmarks needed to be localized. This number varies by the type of pose estimation being done and by the method itself. Table \ref{tab:my_label} lists a few methods for body, face, and hand pose estimation each, as well as the corresponding number of landmarks they used. See Figure \ref{fig:HPE-landmarks} for examples of the same type of pose estimation that vary in the number of the landmarks used. 

\begin{table}[h]
    \begin{tabular}{l|c|c}
Method/Source Name & Type of HPE & Number of Landmarks\\
\hline
HPE OpenCV Github \cite{gb:quanhua} & Body & 18 \\
Lightweight OpenPose Multi-Person HPE
\cite{osokin2018lightweight_openpose} & Body & 18 \\
Whole-Body HPE in the Wild \cite{jin2020wholebody} & Body & 23 \\
BlazePose \cite{website:MP-Body} & Body & 33 \\
\hline
Whole-Body HPE in the Wild \cite{jin2020wholebody} & Face & 68 \\
CNN Facial Landmark Github \cite{gb:bing-cnn-facial} & Face & 68 \\
Facial Keypoint Detection Github \cite{gb:PBB} & Face & 68 \\
MediaPipe Face Mesh \cite{website:mp-facemesh} & Face & 468 \\
\hline
Whole-Body HPE in the Wild \cite{jin2020wholebody} & Hands & 21 per hand \\
MediaPipe Hands \cite{website:mp-hands} & Hands & 21 per hand \\ 
Intro to 2D Hand Pose Estimation \cite{website:rs-olha-hands} & Hands & 21 per hand \\
CNN for 3D Hand Pose Estimation \cite{website:hindawi-cnn-hands} & Hands & 21 per hand \\
\hline
\end{tabular}
    \caption{Standards for Different Types of HPE}
    \label{tab:my_label}
\end{table}

As shown in Table \ref{tab:my_label}, most of the methods for face pose estimation used 68 landmarks because finding algorithms besides MediaPipe Face Mesh that used a number of face landmarks other than 68 proved to be unsuccessful. However, if methods used one of the face datasets for training that does not label its images using 68 landmarks, then there would have been resulting standards that had a number of landmarks other than 68 besides the MediaPipe Face Mesh method that used 468. Similarly, all the methods for hand pose estimation used 21 landmarks per hand because finding algorithms that used a number of hand landmarks other than 21 per hand proved to be unsuccessful as well. If methods used one of the datasets for training that does not label its images using 21 landmarks per hand, then there would have been resulting standards that had a number of landmarks other than 21 per hand. The different types of datasets in HPE are covered in Section \ref{datasets}.

\vspace{8cm}
\subsection{Categorization for HPE}
A way to categorize the different types of pose estimation is based on resolution and the number of landmarks, as this is reflective of the type of pose estimation being done, i.e. hands, face, or body. This approach makes sense due to the properties of these estimations, which will be highlighted when making the categorizations down below.
\begin{itemize}
    \item \underline{First category:} low resolution with up to 30 landmarks
    \begin{itemize}
        \item If there are not many landmarks to be identified, low resolution is sufficient to accomplish this task. This applies to body pose estimation since most cases have around 20 landmarks while some have slightly more than 30, as shown in Table \ref{tab:my_label}. Additionally, the problem of occlusion and complex poses that is prevalent in body pose estimation is handled by a large receptive field and does not require high resolution; thus, low resolution is sufficient enough to identify the landmarks \cite{jin2020wholebody}.
    \end{itemize}
    \item \underline{Second category:} high resolution for dense poses
    \begin{itemize}
        \item  If there are a lot more than 30 landmarks that are both large and dense, higher resolution is required for accurate localization of them. Face pose estimation cases typically have 68 landmarks, and one of them has 468, as shown in Table \ref{tab:my_label}. Similarly, hand pose estimation has 42 total landmarks, or 21 per hand, as also shown in Table \ref{tab:my_label}. In order to accommodate the large quantity and size of their landmarks \cite{jin2020wholebody}, both these types of pose estimation consequently have the property of high resolution, thus making this categorization applicable to them. 
    \end{itemize}
\end{itemize}

\vspace{12cm}
\section{Datasets Used In HPE}
\label{datasets}
Datasets are an important aspect in machine learning. In order for machine learning models to carry out a task, which in this case is HPE, their algorithms must first be trained and then tested to ensure they are correctly interpreting data in order to accomplish the task. This is done through the use of datasets, which consist of training and testing data. There are several of them for each type of HPE, which are described below. Click on the name of each of the datasets to go to their respective main website.
\subsection{Body Datasets}
\begin{itemize}
    \item \href{https://cocodataset.org/#home}{\textbf{COCO}} \cite{website:COCO-overview} \cite{zheng2021deep}: This is the most widely used 2D body dataset, primarily for multi-person HPE. Although it is used for object detection and contains images to help with that, it still has more than 330k images and 200k people labelled with keypoints, up to 17 of them across the whole body, to help with body pose estimation. The first set was released in 2014, but has since been modified. There are 2 versions of the COCO datasets for HPE: 2016 and 2017 COCO keypoints, the difference being the training, validation, and test split.
    \item \href{http://human-pose.mpi-inf.mpg.de/#}{\textbf{MPII}} \cite{andriluka14cvpr} \cite{website:MPII-keypoints} \cite{zheng2021deep}: This 2D body dataset is used primarily for single person pose estimation. It has around 25k images containing 40k individuals with up to 16 manually annotated body joints; this is different from COCO since that has 17 body joints labelled. The images cover 410 different human activities, such as dancing, running, and hunting, and are labelled with that activity. Each image was taken from a YouTube video and was provided both preceding and succeeding frames that were not annotated, which is another difference from COCO. Additionally, advanced annotations like body part occlusions and 3D torso and head orientations were labelled. Made in 2014. 
    \item \href{https://github.com/AIChallenger/AI_Challenger_2017}{\textbf{AI Challenger Human Keypoint Detection}} \cite{Wu_2019} \cite{gb:aichallenger} \cite{zheng2021deep}: This 2D body dataset is the largest one when it comes to 2D HPE. It has over 300k high resolution, annotated images for keypoint detection (14 keypoints per person), and over 600k testing images. All the images were gathered from internet search engines, and, similar to the MPII dataset, focus on daily activities people do in different poses. The difference is they were not taken from YouTube, and it has two additional features to it: attribute based zero-shot recognition, where the machine learns based on descriptions of objects in the image, and image Chinese captioning, where Chinese captions illustrate the relation between objects in the image. Made in 2017. 
    \item \href{https://posetrack.net/}{\textbf{PoseTrack}} \cite{andriluka2018posetrack} \cite{PoseTrack} \cite{zheng2021deep}: This is a 2D video based body dataset that contains around 1356 video sequences, 46k annotated video frames, and 276k body pose annotations. It is primarily used for multi-person body pose estimation where each person has a unique track ID with annotations, which is up to 15 body keypoints. PoseTrack differs from the other body datasets due to its use of videos instead of images and a different number of body keypoints. Made in 2017.
    \item \href{http://vision.imar.ro/human3.6m/}{\textbf{Human3.6M}} \cite{h36m_pami} \cite{zheng2021deep}: This is the most popular and biggest indoor body dataset used for 3D HPE. It contains 3.6 million 3D human poses and uses 3 protocols with different training and testing data splits, as well as 11 professional actors performing 17 activities from 4 different views in an indoor laboratory environment. They are labelled with 24 body keypoints, which is how it differs from the other body datasets, in addition to containing 3D poses instead of 2D. Made in 2014. 
\end{itemize}

\subsection{Face Datasets}
\begin{itemize}
    \item \href{https://ibug.doc.ic.ac.uk/resources/300-W/}{\textbf{300W}} \cite{website:300W-site} \cite{6755925} \cite{khabarlak2021fast}: This 2D face dataset is a collection of several other datasets, including HELEN, AFW, LFPW, and IBUG, that were labelled with 68 landmarks, meaning 300W labels its images using 68 landmarks too. In total, it has around 4000 training images and 600 test images: 300 indoor face images and 300 outdoor in-the-wild images. They all had different shooting conditions, such as lighting and color, emotions, occlusion, face size, face angles, and number of faces present. Made in 2013. 
    \item \href{https://www.tugraz.at/institute/icg/research/team-bischof/lrs/downloads/aflw/}{\textbf{AFLW and related sets}} \cite{koestinger11a} \cite{khabarlak2021fast}: This 2D face dataset has around 25k face images varying in appearance, such as pose, gender, age, ethnicity, and expression. A difference with this set is that it is labelled with only 21 landmarks and has a higher face shooting angle range compared to 300W. There is a version of this dataset that is relabeled with 68 landmarks known as \underline{AFLW-68} but is not used much. Likewise, a dataset known as \underline{MERL-RAV} also relabeled AFLW using 68 landmarks, where each landmark has an extra visibility label, such as visibility, self-occlusion, and occlusion by other objects. Made in 2011.
    \item \href{http://www.vision.caltech.edu/xpburgos/ICCV13/#dataset}{\textbf{COFW and related set}} \cite{6751298} \cite{khabarlak2021fast}: This 2D face dataset is primarily focused on labeling face images that are partially occluded by other objects or the person itself. Consists of around 1.3k training and 507 test images that are labelled with 29 landmarks, which is how it differs from the other face datasets. Similar to AFLW, the test images have also been relabeled into a dataset using 68 landmarks known as \underline{COFW-68}, which can be used to assess landmark detection quality when different datasets are used to train the network. Made in 2013.
    \item \href{https://wywu.github.io/projects/LAB/WFLW.html}{\textbf{WFLW}} \cite{wayne2018lab} \cite{khabarlak2021fast}: This 2D face dataset is one of the most detailed datasets. It consists of a variety of 7.5k training and 2.5k test images; specifically, they have a wide range of emotions, poses, occlusion, blurriness, and lighting conditions. Additionally, they are labelled using 98 densely annotated landmarks, which is also how it differs from the other face datasets. Made in 2018.
\end{itemize}

\subsection{Hand Datasets}
\begin{itemize}
    \item \href{https://sites.google.com/site/qiyeincv/home/bibtex_cvpr2017}{\textbf{BigHand2.2M}} \cite{yuan2017bighand22m} \cite{Yuan_CVPR_2017} \cite{doosti2019hand}: This 3D hand dataset is the biggest hand dataset thus far, as it contains 2.2 million depth images of single hands with 21 annotated keypoints by using 10 subjects. Varying the subjects’ position and arm orientation helped to create a diverse set of view points of the hands. The dataset is broken up into 3 parts: 1.5M frames of schemed poses that cover all the articulations that a human hand can freely adopt; 375k frames of random poses that cover the subjects using their hands to explore the defined pose space; 290k frames of egocentric poses that cover the subjects carrying out 32 extremal poses, which are hand poses where each finger assumes a maximally bent or extended position, combined with random movement. Additionally, since the whole dataset was annotated using kinematic sensors, no objects were held in hands. Made in 2017.
    \item \href{https://handtracker.mpi-inf.mpg.de/projects/GANeratedHands/GANeratedDataset.htm}{\textbf{GANerated Hand Dataset}} \cite{GANeratedHands_CVPR2018} \cite{doosti2019hand}: This 2D and 3D hand dataset contains 330k frames that are synthesized hand shapes and have 21 annotated keypoints. Similar to BigHand2.2M, kinematic sensors were used to capture the hand poses. However, what makes them different is GANerated involved artificial objects being held by hands in order to produce hand occlusion. Made in 2018.
    \item \href{https://jonathantompson.github.io/NYU_Hand_Pose_Dataset.htm}{\textbf{NYU Hand Dataset}} \cite{tompson14tog} \cite{doosti2019hand}: This 3D hand dataset contains over 72k frames in the training set from a single subject and over 8k frames from two different subjects in the test set. What stands out about this dataset compared to the other hand datasets is that it is an RGB-D one captured from a 3rd person view, in addition to labelling the images by using 36 keypoints. Made in 2014. 
    \item \href{http://www.cs.technion.ac.il/~twerd/HandNet/}{\textbf{HandNet Dataset}} \cite{WetzlerBMVC15} \cite{doosti2019hand}: This 3D hand dataset is one of the biggest depth datasets. It contains 202k training frames and 10k test frames. This was all made by using kinematic sensors with 5 male and 5 female subjects in order for the dataset to contain different hand sizes. The images were labelled using 6 keypoints. Made in 2015.
\end{itemize}

\subsection{Whole Body Datasets}
\begin{itemize}
    \item \href{https://github.com/jin-s13/COCO-WholeBody}{\textbf{COCO-WholeBody} \cite{jin2020wholebody}} \cite{jin2020whole}: This 2D whole-body dataset is an extension of the COCO dataset with whole-body annotations, meaning it has manual annotations of landmarks on the entire body, including body, face, hand, and feet, for a total of 133 keypoints (17 for body, 6 for feet, 42 for hands, 68 for face). It has about 130k annotated boxes for each of the two hands and the face, as well as 800k hand keypoints and 4M face keypoints. The set also has 200k images and 250k instances. It is the first dataset with whole-body annotations, as they previously did not exist. Made in 2020.
\end{itemize}
\vspace{5cm}

\section{2D vs 3D HPE}
As shown in Section \ref{datasets}, there are datasets that contain 3D poses. This is because HPE of all types can also be done in 3D and is not limited to just 2D. Listed below are a few papers and sources for body, face, and hand pose estimation in both 2D and 3D, as well as their year of publication and corresponding GitHub repository \cite{gb:HPE-Paper-links} \cite{gb:Hands-Paper-links}. 

\subsection{2D Body Pose Estimation}

\begin{center}
     \begin{tabular}{m{5.4cm} | m{7.4cm} | m{1.5cm}}
Paper/Source Name & GitHub Repository & Publication Year \\
\hline
EvoPose2D: Pushing the Boundaries of 2D Human Pose Estimation using
Neuroevolution \cite{mcnally2020evopose2d} & \url{https://github.com/wmcnally/evopose2d} & 2020 \\
\hline
Efficient Online Multi-Person 2D Pose Tracking 
with Recurrent Spatio-Temporal Affinity Fields \cite{raaj2019efficient} & \url{https://github.com/soulslicer/STAF/tree/staf} & 2019 \\ 
\hline
 Realtime Multi-Person 2D Pose Estimation using Part Affinity Fields \cite{cao2017realtime} & \url{https://github.com/ZheC/Realtime_Multi-Person_Pose_Estimation} & 2017 \\ 
\hline
\end{tabular}
\end{center}
\vspace{0.1cm}

\vspace{0.5cm}
\subsection{2D Face Pose Estimation}
\begin{center}
\begin{savenotes}
   \begin{tabular}{m{5.4cm} | m{7.4cm} | m{1.5cm}}
Paper/Source Name & GitHub Repository & Publication Year \\
\hline
Facial-Keypoint-Detection-Udacity-PPB \cite{gb:PBB} & \url{https://github.com/ParthaPratimBanik/Facial-Keypoint-Detection-Udacity-PPB} & 2021 \\
\hline
High-Resolution Representations for Labeling Pixels and Region\footnote{The model uses 2D face datasets, so this is 2D face pose estimation} \cite{sun2019highresolution} & \url{https://github.com/HRNet/HRNet-Facial-Landmark-Detection} & 2019 \\
\hline
How far are we from solving the 2D \& 3D Face Alignment problem? (and a
dataset of 230,000 3D facial landmarks)\footnote{This paper discusses both 2D and 3D face pose estimation, so it is included in both the 2D and 3D face pose estimation tables} \cite{bulat2017far} & \url{https://github.com/1adrianb/face-alignment} & 2017 \\
\hline
\end{tabular}
\end{savenotes}
\end{center}
\vspace{0.1cm}

\subsection{2D Hand Pose Estimation}
\begin{center}
\begin{savenotes}
    \begin{tabular}{m{5.4cm} | m{7.4cm} | m{1.5cm}}
Paper/Source Name & GitHub Repository & Publication Year\\
\hline
Attention! A Lightweight 2D Hand Pose Estimation Approach \cite{9171866} & \url{https://github.com/nsantavas}\footnote{After clicking on the link, click on the repository titled "Attention-A-Lightweight-2D-Hand-Pose-Estimation-Approach" to access it} & 2021 \\
\hline
Intro to 2D Hand Pose Estimation \cite{website:rs-olha-hands} & \url{https://github.com/OlgaChernytska/2D-Hand-Pose-Estimation-RGB} & 2021 \\
\hline
Nonparametric Structure Regularization Machine for 2D Hand Pose Estimation \cite{chen2020nonparametric} & \url{https://github.com/HowieMa/NSRMhand} & 2020 \\
\hline
\end{tabular}
\end{savenotes}
\end{center}
\vspace{0.1cm}

\vspace{0.5cm}

\subsection{3D Body Pose Estimation}
\begin{center}
  \begin{tabular}{m{5.4cm} | m{7.4cm} | m{1.5cm}}
Paper/Source Name & GitHub Repository & Publication Year\\
\hline
VoxelPose: Towards Multi-Camera 3D Human Pose Estimation in Wild Environment
 \cite{tu2020voxelpose} & \url{https://github.com/microsoft/voxelpose-pytorch} & 2020\\
\hline
Monocular 3D Human Pose Estimation by Generation and Ordinal Ranking
 \cite{sharma2019monocular} & \url{https://github.com/ssfootball04/generative_pose} & 2019 \\
\hline
3D Human Pose Estimation with 2D Marginal Heatmaps \cite{nibali2018margipose} & \url{https://github.com/anibali/margipose} & 2018 \\
\hline
\end{tabular}
\end{center}
\vspace{0.1cm}

\vspace{5.5cm}
\subsection{3D Face Pose Estimation}
\begin{center}
\begin{savenotes}
    \begin{tabular}{m{5.4cm} | m{7.4cm} | m{1.5cm}}
Paper/Source Name & GitHub Repository & Publication Year \\
\hline
Face Landmark Detection With MediaPipe and OpenCV\footnote{The code uses MediaPipe Face Mesh, which uses 468 3D keypoints, so this is 3D face pose estimation \cite{website:MP-Face}} \cite{gb:BakingBrainsFace} & \url{https://github.com/BakingBrains/Face_LandMark_Detection} & 2021 \\
\hline
Multi-view consensus CNN for 3D facial landmark placement \cite{paulsen2018multi} & \url{https://github.com/RasmusRPaulsen/Deep-MVLM} & 2018 \\
\hline
How far are we from solving the 2D \& 3D Face Alignment problem? (and a
dataset of 230,000 3D facial landmarks)\footnote{ This paper discusses both 2D and 3D face pose estimation, so it is included in both the 2D and 3D face pose estimation tables} \cite{bulat2017far} & \url{https://github.com/1adrianb/face-alignment} & 2017 \\
\hline 
\end{tabular}
\end{savenotes}
\end{center}

\vspace{0.1cm}

\vspace{0.5cm}
\subsection{3D Hand Pose Estimation}
\begin{center}
     \begin{tabular}{m{5.4cm} | m{7.4cm} | m{1.5cm}}
Paper/Source Name & GitHub Repository & Publication Year \\
\hline
Active Learning for Bayesian 3D Hand Pose Estimation \cite{caramalau2021active} & \url{https://github.com/razvancaramalau/al_bhpe} & 2021 \\
\hline
HandAugment: A Simple Data Augmentation Method for Depth-Based 3D Hand
Pose Estimation \cite{zhang2020handaugment}
 & \url{https://github.com/wozhangzhaohui/HandAugment} & 2020 \\
\hline
Learning to Estimate 3D Hand Pose from Single RGB Images \cite{zb2017hand} & \url{https://github.com/lmb-freiburg/hand3d} & 2017\\
\hline
\end{tabular}
\end{center}
\vspace{14cm}

\section{Pose Estimation on Images and Videos}
As shown in Section \ref{datasets}, there are datasets that contain videos as part of training and testing the models. This is because there are some algorithms that require video as input, while some require a single image. Section \ref{algo-single-image-input} contains a table listing the names of a few GitHub repositories centered on single image input based on their README.md files, in addition to listing their corresponding links and their year of publication. Similarly, Section \ref{algo-video-input} contains a table that does this for GitHub repositories that have algorithms centered on video input. 

\subsection{Algorithms with Single Image Input}
\label{algo-single-image-input}
\begin{center}
\begin{savenotes}
     \begin{tabular}{m{7cm} | m{6.5cm} | m{1.5cm}}
GitHub Repository Name (Based on README.md File) & Link to GitHub Repository & Publication Year\\
\hline
MeTRAbs Absolute 3D Human Pose Estimator \cite{Sarandi20TBIOM} & \url{https://github.com/isarandi/metrabs} & 2020 \\
\hline
RootNet for 3D Multi-person Pose Estimation on Single RGB Image \cite{Moon_2019_ICCV_3DMPPE} & \url{https://github.com/mks0601/3DMPPE_ROOTNET_RELEASE} & 2019 \\
\hline
IBM Developer Model Asset Exchange: Human Pose Estimator \cite{gb:IBM} & \url{https://github.com/IBM/MAX-Human-Pose-Estimator} & 2018 \\
\hline
human-pose-estimation-opencv\footnote{Specify image input when running the python script using the command line} \cite{gb:quanhua} & \url{https://github.com/quanhua92/human-pose-estimation-opencv} & 2018 \\
\hline
Lifting from the Deep \cite{Tome_2017_CVPR} & \url{https://github.com/DenisTome/Lifting-from-the-Deep-release} & 2017 \\
\hline
\end{tabular}
\end{savenotes}
\end{center}

\subsection{Algorithms with Video Input}
\label{algo-video-input}
\begin{center}
\begin{savenotes}
     \begin{tabular}{m{7cm} | m{6.5cm} | m{1.5cm}}
GitHub Repository Name (Based on README.md File) & Link to GitHub Repository & Publication Year\\
\hline
Pose\_estimation\footnote{The .py file uses the VideoCapture function, which is used for video input } \cite{gb:BakingBrains} & \url{https://github.com/BakingBrains/Pose_estimation} & 2021 \\
\hline
MediaPipe Pose Estimation Project\footnote{Specify video input when running the python script using the command line \cite{website:MP-OpenCV-HPE-Video}} \cite{gb:arthurfortes} & \url{https://github.com/arthurfortes/pose_estimation} & 2021 \\
\hline
GAST-Net for 3D Human Pose Estimation in Video \cite{liu2020a} & \url{https://github.com/fabro66/GAST-Net-3DPoseEstimation} & 2020 \\
\hline
3D HPE in video with temporal convolutions and semi-supervised training \cite{pavllo:videopose3d:2019} & \url{https://github.com/facebookresearch/VideoPose3D} & 2019 \\
\hline
Detect And Track: Efficient Pose Estimation in Videos \cite{girdhar2018detecttrack} & \url{https://github.com/facebookresearch/DetectAndTrack/} & 2018 \\
\hline
\end{tabular}
\end{savenotes}
\end{center}

\bibliographystyle{unsrt}
\bibliography{citations}

\begin{thebibliography}{10}

\bibitem{website:nanonets-2019}
Sudharshan~Chandra Babu.
\newblock A 2019 guide to human pose estimation with deep learning.
\newblock \url{https://nanonets.com/blog/human-pose-estimation-2d-guide/},
  2019.

\bibitem{website:mp-hands}
Google.
\newblock Mediapipe hands.
\newblock \url{https://google.github.io/mediapipe/solutions/hands.html}, 2020.

\bibitem{gb:adrianb}
1adrianb.
\newblock face-alignment.
\newblock
  \url{https://github.com/1adrianb/face-alignment/blob/master/docs/images/face-alignment-adrian.gif},
  2017.

\bibitem{FMP}
Yi~Yang and Deva Ramanan.
\newblock Articulated human detection with flexible mixtures of parts.
\newblock {\em IEEE Transactions on Pattern Analysis and Machine Intelligence},
  35(12):2878--2890, 2013.

\bibitem{website:neuralet-DL-Classics}
Human pose estimation with deep learning (part i).
\newblock
  \url{https://neuralet.com/article/human-pose-estimation-with-deep-learning-part-i/}.

\bibitem{wei2016convolutional}
Shih-En Wei, Varun Ramakrishna, Takeo Kanade, and Yaser Sheikh.
\newblock Convolutional pose machines, 2016.

\bibitem{website:review-cpm}
Sik-Ho Tsang.
\newblock Review: Cpm — convolutional pose machines (human pose estimation).
\newblock
  \url{https://sh-tsang.medium.com/review-cpm-convolutional-pose-machines-human-pose-estimation-224cfeb70aac},
  2019.

\bibitem{website:OKS-PDJ}
Alexander Stasiuk.
\newblock Pose estimation. metrics.
\newblock
  \url{https://alexander-stasiuk.medium.com/pose-estimation-metrics-844c07ba0a78},
  2020.

\bibitem{zheng2021deep}
Ce~Zheng, Wenhan Wu, Taojiannan Yang, Sijie Zhu, Chen Chen, Ruixu Liu, Ju~Shen,
  Nasser Kehtarnavaz, and Mubarak Shah.
\newblock Deep learning-based human pose estimation: A survey, 2021.

\bibitem{website:IOU}
5 object detection evaluation metrics that data scientists should know.
\newblock
  \url{https://analyticsindiamag.com/5-object-detection-evaluation-metrics-that-data-scientists-should-know/},
  2020.

\bibitem{website:MAR/MAP/AR/AP}
How the compute accuracy for object detection tool works.
\newblock
  \url{https://pro.arcgis.com/en/pro-app/latest/tool-reference/image-analyst/how-compute-accuracy-for-object-detection-works.htm}.

\bibitem{website2:MAR/MAP/AR/AP}
An introduction to evaluation metrics for object detection.
\newblock
  \url{https://blog.zenggyu.com/en/post/2018-12-16/an-introduction-to-evaluation-metrics-for-object-detection/},
  2018.

\bibitem{website:COCO-metrics}
Tsung-Yi Lin, Genevieve Patterson, Matteo~R. Ronchi, Yin Cui, Michael Maire,
  Serge Belongie, Lubomir Bourdev, Ross Girshick, James Hays, Pietro Perona,
  Deva Ramanan, C.~Lawrence Zitnick, and Piotr Dollár.
\newblock Coco dataset.
\newblock \url{https://cocodataset.org/#keypoints-eval}, 2014.

\bibitem{gb:quanhua}
quanhua92.
\newblock human-pose-estimation-opencv.
\newblock \url{https://github.com/quanhua92/human-pose-estimation-opencv}, June
  2018.

\bibitem{osokin2018lightweight_openpose}
Daniil Osokin.
\newblock Real-time 2d multi-person pose estimation on cpu: Lightweight
  openpose.
\newblock In {\em arXiv preprint arXiv:1811.12004}, 2018.

\bibitem{jin2020wholebody}
Sheng Jin, Lumin Xu, Jin Xu, Can Wang, Wentao Liu, Chen Qian, Wanli Ouyang, and
  Ping Luo.
\newblock Whole-body human pose estimation in the wild, 2020.

\bibitem{website:MP-Body}
Valentin Bazarevsky and Ivan Grishchenko.
\newblock On-device, real-time body pose tracking with mediapipe blazepose.
\newblock
  \url{https://ai.googleblog.com/2020/08/on-device-real-time-body-pose-tracking.html},
  August 2020.

\bibitem{gb:bing-cnn-facial}
yinguobing.
\newblock cnn-facial-landmark.
\newblock \url{https://github.com/yinguobing/cnn-facial-landmark}, March 2021.

\bibitem{gb:PBB}
ParthaPratimBanik.
\newblock Facial-keypoint-detection-udacity-pbb.
\newblock
  \url{https://github.com/ParthaPratimBanik/Facial-Keypoint-Detection-Udacity-PPB},
  February 2021.

\bibitem{website:mp-facemesh}
Google.
\newblock Mediapipe face mesh.
\newblock \url{https://google.github.io/mediapipe/solutions/face_mesh}, 2020.

\bibitem{website:rs-olha-hands}
Olga Chernytska.
\newblock Gentle introduction to 2d hand pose estimation: Approach explained.
\newblock
  \url{https://notrocketscience.blog/gentle-introduction-to-2d-hand-pose-estimation-lets-code-it/},
  April 2021.

\bibitem{website:hindawi-cnn-hands}
Shiming Dai, Wei Liu, Wenji Yang, Lili Fan, and Jihao Zhang.
\newblock Cascaded hierarchical cnn for rgb-based 3d hand pose estimation.
\newblock \url{https://www.hindawi.com/journals/mpe/2020/8432840/}, July 2020.

\bibitem{website:COCO-overview}
Tsung-Yi Lin, Genevieve Patterson, Matteo~R. Ronchi, Yin Cui, Michael Maire,
  Serge Belongie, Lubomir Bourdev, Ross Girshick, James Hays, Pietro Perona,
  Deva Ramanan, C.~Lawrence Zitnick, and Piotr Dollár.
\newblock Coco dataset.
\newblock \url{https://cocodataset.org/#home}, 2014.

\bibitem{andriluka14cvpr}
Mykhaylo Andriluka, Leonid Pishchulin, Peter Gehler, and Bernt Schiele.
\newblock 2d human pose estimation: New benchmark and state of the art
  analysis.
\newblock In {\em IEEE Conference on Computer Vision and Pattern Recognition
  (CVPR)}, June 2014.

\bibitem{website:MPII-keypoints}
Mpii (mpii human pose).
\newblock \url{https://paperswithcode.com/dataset/mpii}.

\bibitem{Wu_2019}
Jiahong Wu, He~Zheng, Bo~Zhao, Yixin Li, Baoming Yan, Rui Liang, Wenjia Wang,
  Shipei Zhou, Guosen Lin, Yanwei Fu, and et~al.
\newblock Large-scale datasets for going deeper in image understanding.
\newblock {\em 2019 IEEE International Conference on Multimedia and Expo
  (ICME)}, Jul 2019.

\bibitem{gb:aichallenger}
AIChallenger.
\newblock Ai\_challenger\_2017.
\newblock \url{https://github.com/AIChallenger/AI_Challenger_2017}, 2017.

\bibitem{andriluka2018posetrack}
Mykhaylo Andriluka, Umar Iqbal, Eldar Insafutdinov, Leonid Pishchulin, Anton
  Milan, Juergen Gall, and Bernt Schiele.
\newblock Posetrack: A benchmark for human pose estimation and tracking, 2018.

\bibitem{PoseTrack}
M.~Andriluka, U.~Iqbal, E.~Ensafutdinov, L.~Pishchulin, A.~Milan, J.~Gall, and
  Schiele B.
\newblock Pose{T}rack: {A} benchmark for human pose estimation and tracking.
\newblock In {\em CVPR}, 2018.

\bibitem{h36m_pami}
Catalin Ionescu, Dragos Papava, Vlad Olaru, and Cristian Sminchisescu.
\newblock Human3.6m: Large scale datasets and predictive methods for 3d human
  sensing in natural environments.
\newblock {\em IEEE Transactions on Pattern Analysis and Machine Intelligence},
  2014.

\bibitem{website:300W-site}
300 faces in-the-wild challenge (300-w), iccv 2013.
\newblock \url{https://ibug.doc.ic.ac.uk/resources/300-W/}.

\bibitem{6755925}
Christos Sagonas, Georgios Tzimiropoulos, Stefanos Zafeiriou, and Maja Pantic.
\newblock 300 faces in-the-wild challenge: The first facial landmark
  localization challenge.
\newblock In {\em 2013 IEEE International Conference on Computer Vision
  Workshops}, pages 397--403, 2013.

\bibitem{khabarlak2021fast}
Kostiantyn Khabarlak and Larysa Koriashkina.
\newblock Fast facial landmark detection and applications: A survey, 2021.

\bibitem{koestinger11a}
Peter M.~Roth Martin~Koestinger, Paul~Wohlhart and Horst Bischof.
\newblock {Annotated Facial Landmarks in the Wild: A Large-scale, Real-world
  Database for Facial Landmark Localization}.
\newblock In {\em {Proc. First IEEE International Workshop on Benchmarking
  Facial Image Analysis Technologies}}, 2011.

\bibitem{6751298}
Xavier~P. Burgos-Artizzu, Pietro Perona, and Piotr Dollár.
\newblock Robust face landmark estimation under occlusion.
\newblock In {\em 2013 IEEE International Conference on Computer Vision}, pages
  1513--1520, December 2013.

\bibitem{wayne2018lab}
Wayne Wu, Chen Qian, Shuo Yang, Quan Wang, Yici Cai, and Qiang Zhou.
\newblock Look at boundary: A boundary-aware face alignment algorithm.
\newblock In {\em CVPR}, June 2018.

\bibitem{yuan2017bighand22m}
Shanxin Yuan, Qi~Ye, Bjorn Stenger, Siddhant Jain, and Tae-Kyun Kim.
\newblock Bighand2.2m benchmark: Hand pose dataset and state of the art
  analysis, 2017.

\bibitem{Yuan_CVPR_2017}
Shanxin Yuan, Qi~Ye, Bj\"{o}rn Stenger, Siddhant Jain, and Tae-Kyun Kim.
\newblock Bighand2.2m benchmark: Hand pose dataset and state of the art
  analysis.
\newblock In {\em CVPR}, 2017.

\bibitem{doosti2019hand}
Bardia Doosti.
\newblock Hand pose estimation: A survey, 2019.

\bibitem{GANeratedHands_CVPR2018}
Franziska Mueller, Florian Bernard, Oleksandr Sotnychenko, Dushyant Mehta,
  Srinath Sridhar, Dan Casas, and Christian Theobalt.
\newblock Ganerated hands for real-time 3d hand tracking from monocular rgb.
\newblock In {\em Proceedings of Computer Vision and Pattern Recognition
  ({CVPR})}, June 2018.

\bibitem{tompson14tog}
Jonathan Tompson, Murphy Stein, Yann Lecun, and Ken Perlin.
\newblock Real-time continuous pose recovery of human hands using convolutional
  networks.
\newblock {\em ACM Transactions on Graphics}, 33, August 2014.

\bibitem{WetzlerBMVC15}
Aaron Wetzler, Ron Slossberg, and Ron Kimmel.
\newblock Rule of thumb: Deep derotation for improved fingertip detection.
\newblock In Mark W.~Jones Xianghua~Xie and Gary K.~L. Tam, editors, {\em
  Proceedings of the British Machine Vision Conference (BMVC)}, pages
  33.1--33.12. BMVA Press, September 2015.

\bibitem{jin2020whole}
Sheng Jin, Lumin Xu, Jin Xu, Can Wang, Wentao Liu, Chen Qian, Wanli Ouyang, and
  Ping Luo.
\newblock Whole-body human pose estimation in the wild.
\newblock In {\em Proceedings of the European Conference on Computer Vision
  (ECCV)}, 2020.

\bibitem{gb:HPE-Paper-links}
wangzheallen.
\newblock awesome-human-pose-estimation.
\newblock
  \url{https://github.com/wangzheallen/awesome-human-pose-estimation#2d-pose-estimation},
  August 2020.

\bibitem{gb:Hands-Paper-links}
xinghaochen.
\newblock awesome-hand-pose-estimation.
\newblock \url{https://github.com/xinghaochen/awesome-hand-pose-estimation},
  2021.

\bibitem{mcnally2020evopose2d}
William McNally, Kanav Vats, Alexander Wong, and John McPhee.
\newblock Evopose2d: Pushing the boundaries of 2d human pose estimation using
  neuroevolution, 2020.

\bibitem{raaj2019efficient}
Yaadhav Raaj, Haroon Idrees, Gines Hidalgo, and Yaser Sheikh.
\newblock Efficient online multi-person 2d pose tracking with recurrent
  spatio-temporal affinity fields, 2019.

\bibitem{cao2017realtime}
Zhe Cao, Tomas Simon, Shih-En Wei, and Yaser Sheikh.
\newblock Realtime multi-person 2d pose estimation using part affinity fields,
  2017.

\bibitem{sun2019highresolution}
Ke~Sun, Yang Zhao, Borui Jiang, Tianheng Cheng, Bin Xiao, Dong Liu, Yadong Mu,
  Xinggang Wang, Wenyu Liu, and Jingdong Wang.
\newblock High-resolution representations for labeling pixels and regions,
  2019.

\bibitem{bulat2017far}
Adrian Bulat and Georgios Tzimiropoulos.
\newblock How far are we from solving the 2d \& 3d face alignment problem? (and
  a dataset of 230,000 3d facial landmarks).
\newblock In {\em International Conference on Computer Vision}, 2017.

\bibitem{9171866}
Nicholas Santavas, Ioannis Kansizoglou, Loukas Bampis, Evangelos Karakasis, and
  Antonios Gasteratos.
\newblock Attention! a lightweight 2d hand pose estimation approach.
\newblock {\em IEEE Sensors Journal}, 21(10):11488--11496, 2021.

\bibitem{chen2020nonparametric}
Yifei Chen, Haoyu Ma, Deying Kong, Xiangyi Yan, Jianbao Wu, Wei Fan, and
  Xiaohui Xie.
\newblock Nonparametric structure regularization machine for 2d hand pose
  estimation.
\newblock In {\em The IEEE Winter Conference on Applications of Computer
  Vision}, pages 381--390, 2020.

\bibitem{tu2020voxelpose}
Hanyue Tu, Chunyu Wang, and Wenjun Zeng.
\newblock Voxelpose: Towards multi-camera 3d human pose estimation in wild
  environment, 2020.

\bibitem{sharma2019monocular}
Saurabh Sharma, Pavan~Teja Varigonda, Prashast Bindal, Abhishek Sharma, and
  Arjun Jain.
\newblock Monocular 3d human pose estimation by generation and ordinal ranking,
  2019.

\bibitem{nibali2018margipose}
Aiden Nibali, Zhen He, Stuart Morgan, and Luke Prendergast.
\newblock 3d human pose estimation with 2d marginal heatmaps.
\newblock {\em arXiv preprint arXiv:1806.01484}, 2018.

\bibitem{website:MP-Face}
Shakhadri313.
\newblock Facial landmark detection simplified with opencv.
\newblock
  \url{https://www.analyticsvidhya.com/blog/2021/07/facial-landmark-detection-simplified-with-opencv/},
  July 2021.

\bibitem{gb:BakingBrainsFace}
BakingBrains.
\newblock Face\_landmark\_detection.
\newblock \url{https://github.com/BakingBrains/Face_LandMark_Detection}, June
  2021.

\bibitem{paulsen2018multi}
Rasmus~R Paulsen, Kristine~Aavild Juhl, Thilde~Marie Haspang, Thomas Hansen,
  Melanie Ganz, and Gudmundur Einarsson.
\newblock Multi-view consensus cnn for 3d facial landmark placement.
\newblock In {\em Asian Conference on Computer Vision}, pages 706--719.
  Springer, 2018.

\bibitem{caramalau2021active}
Razvan Caramalau, Binod Bhattarai, and Tae-Kyun Kim.
\newblock Active learning for bayesian 3d hand pose estimation, 2021.

\bibitem{zhang2020handaugment}
Zhaohui Zhang, Shipeng Xie, Mingxiu Chen, and Haichao Zhu.
\newblock Handaugment: A simple data augmentation method for depth-based 3d
  hand pose estimation, 2020.

\bibitem{zb2017hand}
Christian Zimmermann and Thomas Brox.
\newblock Learning to estimate 3d hand pose from single rgb images.
\newblock In {\em IEEE International Conference on Computer Vision (ICCV)},
  2017.
\newblock https://arxiv.org/abs/1705.01389.

\bibitem{Sarandi20TBIOM}
Istv\'an S\'ar\'andi, Timm Linder, Kai~O. Arras, and Bastian Leibe.
\newblock {MeTRAbs:} metric-scale truncation-robust heatmaps for absolute 3{D}
  human pose estimation.
\newblock {\em IEEE Transactions on Biometrics, Behavior, and Identity
  Science}, 2020.
\newblock in press.

\bibitem{Moon_2019_ICCV_3DMPPE}
Gyeongsik Moon, Juyong Chang, and Kyoung~Mu Lee.
\newblock Camera distance-aware top-down approach for 3d multi-person pose
  estimation from a single rgb image.
\newblock In {\em The IEEE Conference on International Conference on Computer
  Vision (ICCV)}, 2019.

\bibitem{gb:IBM}
IBM.
\newblock Max-human-pose-estimator.
\newblock \url{https://github.com/IBM/MAX-Human-Pose-Estimator}, December 2020.

\bibitem{Tome_2017_CVPR}
Denis Tome, Chris Russell, and Lourdes Agapito.
\newblock Lifting from the deep: Convolutional 3d pose estimation from a single
  image.
\newblock In {\em The IEEE Conference on Computer Vision and Pattern
  Recognition (CVPR)}, July 2017.

\bibitem{gb:BakingBrains}
BakingBrains.
\newblock Pose\_estimation.
\newblock \url{https://github.com/BakingBrains/Pose_estimation}, June 2021.

\bibitem{website:MP-OpenCV-HPE-Video}
Arthur Fortes.
\newblock Deep learning based human pose estimation using opencv and mediapipe.
\newblock
  \url{https://medium.com/nerd-for-tech/deep-learning-based-human-pose-estimation-using-opencv-and-mediapipe-d0be7a834076},
  May 2021.

\bibitem{gb:arthurfortes}
arthurfortes.
\newblock pose\_estimation.
\newblock \url{https://github.com/arthurfortes/pose_estimation}, May 2021.

\bibitem{liu2020a}
Junfa Liu, Juan Rojas, Zhijun Liang, Yihui Li, and Yisheng Guan.
\newblock A graph attention spatio-temporal convolutional networks for 3d human
  pose estimation in video.
\newblock {\em arXiv preprint arXiv:2003.14179}, 2020.

\bibitem{pavllo:videopose3d:2019}
Dario Pavllo, Christoph Feichtenhofer, David Grangier, and Michael Auli.
\newblock 3d human pose estimation in video with temporal convolutions and
  semi-supervised training.
\newblock In {\em Conference on Computer Vision and Pattern Recognition
  (CVPR)}, 2019.

\bibitem{girdhar2018detecttrack}
Rohit Girdhar, Georgia Gkioxari, Lorenzo Torresani, Manohar Paluri, and
  Du~Tran.
\newblock {Detect-and-Track: Efficient Pose Estimation in Videos}.
\newblock In {\em CVPR}, 2018.

\end{thebibliography}
\end{document}